\documentclass[pdflatex,sn-mathphys-num]{sn-jnl}% Math and Physical Sciences Numbered Reference Style
\usepackage{silence}
\usepackage{graphicx}%
\usepackage{multirow}%
\usepackage{amsmath,amssymb,amsfonts}%
\usepackage{amsthm}%
\usepackage{mathrsfs}%
\usepackage[title]{appendix}%
\usepackage{xcolor}%
\usepackage{textcomp}%
\usepackage{manyfoot}%
\usepackage{booktabs}%
\usepackage{algorithm}%
\usepackage{algorithmicx}%
\usepackage{algpseudocode}%
\usepackage{listings}%
\theoremstyle{thmstyleone}%
\theoremstyle{thmstyletwo}%

\theoremstyle{thmstylethree}%
\makeatletter
\@ifpackageloaded{hyperref}{%
  \hypersetup{bookmarksdepth=subsubsection}
}{}
\makeatother

\begin{document}

\title[LightTeaNet: A Weakly Supervised Lightweight CNN for Multi-Label Tea Leaf Disease Detection and Localization]{LightTeaNet: A Weakly Supervised Lightweight CNN for Multi-Label Tea Leaf Disease Detection and Localization}


\author*[1]{\fnm{Naif Haider} \sur{Chowdhury}}\email{cse\_2132020032@lus.ac.bd}

\author[1]{\fnm{Md} \sur{Rahim}}\email{cse\_2132020095@lus.ac.bd}
\equalcont{These authors contributed equally to this work.}

\author[1]{\fnm{Syed Farhan} \sur{Hasan}}\email{cse\_2132020061@lus.ac.bd}
\equalcont{These authors contributed equally to this work.}

\author[1]{\fnm{Murad} \sur{Hasan}}\email{cse\_2132020075@lus.ac.bd}
\equalcont{These authors contributed equally to this work.}

\author[1]{\fnm{Prithwiraj} \sur{Bhattacharjee}}\email{prithwiraj\_cse@lus.ac.bd}
\equalcont{These authors contributed equally to this work.}

\affil*[1]{\orgdiv{Department of Computer Science \& Engineering}, \orgname{Leading University}, \orgaddress{\street{Kamal Bazar}, \city{Sylhet}, \postcode{3112}, \state{Sylhet}, \country{Bangladesh}}}


\abstract{Tea is known as an important crop in many parts of South and Southeast Asia, yet the production of tea is still hampered by the multiple diseases that decrease the quantity and quality. Traditional methods of inspection, which are manual, are not consistent, labor-intensive, and depend on extensive monitoring. This paper introduces a lightweight convolutional neural network (CNN) designed for weakly supervised multi-label classification and disease localization in tea leaves called LightTeaNet. LightTeaNet learns directly from image-level labels and employs Class Activation Mapping (CAM) to localize disease-affected regions automatically, unlike conventional object detection models such as YOLO, which require extensive bounding box annotations.  For Parameter efficiency, the network integrates Depthwise Separable Convolutions, and for enhanced feature discrimination, it integrates Channel Attention. LightTeaNet has achieved a Precision of 0.9615, a Recall of 0.8772, and an F1-score of 0.9179, while it shows mAP@0.50=0.1810 without any manual annotations, which delivers a competitive localization performance in the experimental results. These results validate the model as an interpretable as well as a resource-efficient framework for intelligent disease monitoring in agriculture.}




\maketitle

\section{Introduction}\label{sec:introduction}

The rise of deep learning has changed how agricultural problems are addressed, especially in the field of plant disease 
detection. The healthy production of tea has become important since it is a vital economic crop in many Asian countries. 
Manual methods of inspection are time-consuming, inconsistent, and depend heavily on expert knowledge. This research 
introduces a new approach that leverages artificial intelligence to detect and localize tea leaf diseases more efficiently and 
accurately.

\subsection{Background and Motivation}\label{subsec:background_and_motivation}

The continuous involvement of Artificial Intelligence (AI) and Deep Learning (DL) technologies in agriculture has revolutionized the practices of disease monitoring and crop management \cite{bib32,bib33}. Tea production plays a vital role in regional economies, especially in South and Southeast Asia. But tea leaves are vulnerable to various diseases, for example, Brown Blight, Gray Blight, and Red Spider infestations that can cause huge production losses of tea \cite{bib9,bib14,bib22,bib34}. Therefore, accurate and effective disease detection is needed for the cultivation of tea.

\subsection{Challenges in Existing Approaches}\label{subsec:challenges_in_existing_approaches}

In conventional disease recognition systems, they are primarily focused on single-label classification or fully supervised object 
detection frameworks. Even if there is strong localization accuracy, object detectors like the YOLO depend on manual bounding 
box annotations that are not practical in the case of large datasets. Also, accurate annotation is subjective and prone to errors 
since there is a possibility of visual overlap and subtle inter-disease variations in tea leaves.

\subsection{Research Objective}\label{subsec:research_objective}

As the challenges have been addressed, an architecture that can perform multi-label classification and visual localization of tea leaf diseases is proposed in this study, called LightTeaNet, which is a weakly supervised and annotation-free CNN architecture. The design of the model prioritizes efficiency, interpretability, and ease of deployment. By using CAM-based activation maps that can highlight diseased regions directly from image-level labels, LightTeaNet removes the need for bounding box data and reduces heavy manual labor.

\subsection{Contributions}\label{subsec:contributions}

The key contributions of this research paper are mentioned below:
\begin{itemize}
\item An annotation-free disease localization pipeline that bridges the gap between inexpensive image-level labels and the data-intensive requirements of fully supervised object detection models, enabling effective object detection training without manual bounding box annotations.
Convolutions and Channel Attention to enhance performance.
\item By using Class Activation Mapping (CAM) for weakly supervised localization, without bounding boxes, disease regions can be visualized and interpreted.
\item A comprehensive comparative evaluation using automatically generated pseudoannotations from LightTeaNet's CAMs to train and compare against a broad range of state-of-the-art object detection models, including YOLO variants (YOLOv8n, YOLOv8m, YOLO11n, YOLO12s), Transformer-based detectors (RFDETR, DETR), Two-stage detectors (Faster R-CNN), and Single-stage detectors (RetinaNet).
\item On the TeaLeafBD dataset, State-of-the-art classification outcomes using competitive localization performance under weak supervision.
\end{itemize}

Overall, this study seeks to balance between accuracy and practicality in tea leaf disease detection. It combines lightweight 
architecture, weak supervision, and interpretable visualization through Class Activation Mapping (CAM). The proposed 
LightTeaNet highlights that deep learning can make annotation much easier while decreasing the computational cost, also 
delivering reliable outcomes suitable for real-world agricultural use.

\section{Related Work}\label{sec:related_work}

Several studies have explored deep learning applications in plant disease detection, focusing on improving accuracy, interpretability, and efficiency. Researchers have worked on multi-label classification, data augmentation, and attention mechanisms to make models more robust and generalizable. This section reviews key works that inspired our approach and helped shape the development of LightTeaNet. Among them Liu et al.\ \cite{bib1} found the concept of extreme multi-label classification, which has established the foundation for multi-label image problems. Perez and Wang \cite{bib2} showed that model generalization can be improved by data augmentation, while Maxwell et al.\ \cite{bib3} provided deep learning architectures for multi-label classification.

Woo et al.\ \cite{bib4} influenced modern attention-based CNNs by developing the Convolutional Block Attention Module (CBAM). Selvaraju et al.\ \cite{bib6} proposed Grad-CAM, a key explainability technique, while Shorten and Khoshgoftaar \cite{bib7} conducted a survey on data augmentation practices. Later, Moore \cite{bib8} introduced Class Activation Mapping (CAM), which made it possible to visually locate model attention areas. In accuracy, Bao et al.\ \cite{bib9} have achieved high accuracy in plant pathology by implementing an AX-RetinaNet for tea leaf disease detection. Prasetyo et al.\ \cite{bib10} developed depthwise separable convolutions that enable lightweight CNN architectures. Nanni et al.\ \cite{bib11} emphasized input normalization for stable training. Alomar et al.\ \cite{bib12} further explored robustness and how data augmentation can have a huge impact on it. Yang et al.\ \cite{bib13} applied tea leaf disease detection by deep learning, while Zilliz Learn \cite{bib15} clarified that, for visual interpretation, CAM is very useful. Tarekegn et al.\ \cite{bib16} summarized multi-label problem solutions from deep learning approaches. On the other hand, Ma et al.\ \cite{bib17} proposed adaptive attention modules, and DigitalOcean Tutorial \cite{bib18} introduced a concept integrated into our model's attention design called squeeze-and-excitation (SE) mechanisms. Further, DigitalOcean Tutorial \cite{bib19} reviewed the use of attention mechanisms in vision systems. Yang et al.\ \cite{bib13} presented automated augmentation surveys, while Balasundaram et al.\ \cite{bib22} combined segmentation and classification for tea leaf disease detection using deep CNNs.

Zhang et al.\ \cite{bib23} worked on enhancing convergence stability and then introduced the WuC-Adam optimizer. Liu et al.\ \cite{bib23} analyzed how augmentation affects explainability. Wu et al.\ \cite{bib25} proposed an attention-CNN hybrid for agricultural disease recognition. Fayyaz et al.\ \cite{bib26} systematically reviewed Grad-CAM techniques. Nyckel Guide \cite{bib27} differentiated between multi-class and multi-label tasks, while A Data Odyssey \cite{bib28} explained the CAM generation step-by-step. Wang et al.\ \cite{bib29} explored deep active learning for multi-label systems. Feng et al.\ \cite{bib30} analyzed hybrid attention mechanisms. Vishwakarma \cite{bib31} provided a practical guide for Grad-CAM visualization. Rahat et al.\ \cite{bib32} demonstrated the application of CNNs in precision agriculture, UnitX Labs \cite{bib33} discussed industrial attention-based vision systems, and Dipty et al.\ \cite{bib34} proposed a real-time Android-based tea disease detection model called TeaNet8. Towards Data Science \cite{bib35} presented the MobileNetV1 architecture, central to modern lightweight CNNs. Alam et al.\ \cite{bib36} introduced teaLeafBD, a large annotated dataset of healthy and diseased tea leaves, providing an essential foundation for tea disease research. Das et al.\ \cite{bib41} applied edge-ready tea leaf disease classification with lightweight CNNs.

In summary, previous research has made strong progress in image classification, attention mechanisms, and explainable AI. However, most existing methods depend on manual annotations or focus on single-label classification. To overcome these gaps, our proposed LightTeaNet introduces a lightweight and weakly supervised approach that combines Depthwise Separable Convolutions, Channel Attention, and Class Activation Mapping (CAM) for efficient and interpretable tea leaf disease detection.

\section{Methodology}\label{sec:methodology}

This chapter explains the overall process followed in developing and evaluating our proposed model, LightTeaNet. It starts with dataset preparation and preprocessing to model training, comparison, and performance evaluation. This method was designed to ensure a fair comparison between the existing model and our custom CNN, while keeping the workflow simple and reproducible.

\subsection{Overview}\label{subsec:overview}

The proposed workflow for the identification and localization of tea leaf diseases involves several cycles of procedures, starting with data collection and preprocessing, and continuing through model training and evaluation. The entire pipeline contains several YOLO models for comparison, as well as the proposed custom CNN (LightTeaNet).

\subsection{Flow Diagram of the Full Process}\label{subsec:flow_diagram_of_the_full_process}

\begin{figure}[ht]
\centering
\includegraphics[width=1.0\textwidth]{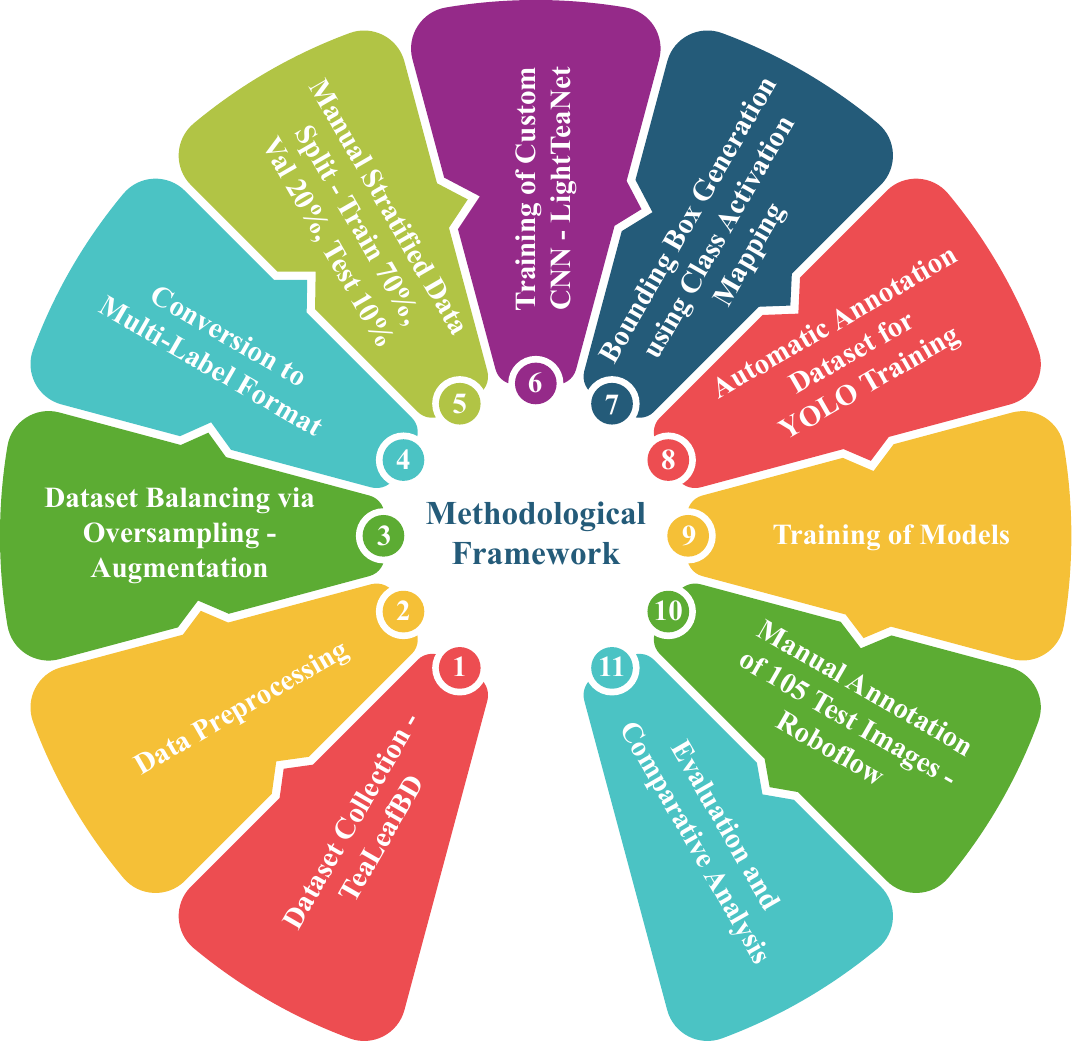}
\caption{Methodological framework from start to end.}\label{fig:figure_1}
\end{figure}

Figure~\ref{fig:figure_1} illustrates the entire methodological framework from start to end.

\subsection{Dataset Preparation}\label{subsec:dataset_preparation}

\subsubsection{Dataset Collection}\label{subsubsec:dataset_collection}

The dataset used in this study, named TeaLeafBD \cite{bib36}, contains seven disease classes: Brown Blight, Gray Blight, Green Mirid Bug, Red Spider, Helopeltis, Tea Algal Leaf Spot, and Healthy. Each image filename corresponds to its associated disease label. Most images contain a single tea leaf centred on a white background, although some images include multiple leaves.

\subsubsection{Dataset Balancing via Augmentation}\label{subsubsec:dataset_balancing_via_augmentation}

There was a class imbalance in the dataset. Data augmentation techniques such as random rotation, flipping both horizontally and vertically, brightness modification, and random cropping were used to oversample in order to solve this problem \cite{bib2,bib5,bib7,bib12,bib13,bib20}. This procedure balanced class distribution and increased data diversity in order to improve model generalization.

\subsubsection{Multi-Label Dataset Conversion}\label{subsubsec:multi-label_dataset_conversion}

As the initial dataset was single-labelled, it was converted to a multi-label representation to allow the model to predict multiple diseases per leaf. Each image was represented as a 7-dimensional binary vector, where 1 indicated the presence of the disease, and 0 indicated its absence.

\subsubsection{Manual Stratified Splitting}\label{subsubsec:manual_stratified_splitting}

A manual stratified splitting technique was used to divide the dataset and maintain a balanced proportion of disease classes throughout all subsets. The data were split into 70\% for training, 20\% for validation, and 10\% for testing. Duplicate images were carefully removed so that no sample appeared in more than one subset. For the sake of experimental consistency and reproducibility, the final split data were saved in separate CSV files.

\subsection{YOLO Models}\label{subsec:yolo_models}

\subsubsection{Architectural Design}\label{subsubsec:architectural_design_yolo}

Training parameters were the same across all YOLO models: 640$\times$640 images, 16 batches, 300 epochs, and 10 early-stopping patience. YOLOv8n pretrained weights were used to initialize them, and resume mode was enabled to allow continuation from previous checkpoints. Importantly, LightTeaNet's Class Activation Mapping (CAM) was used to generate the training data, which included automatically generated bounding boxes.

\subsubsection{Architecture Diagram}\label{subsubsec:architecture_diagram_yolo}

\begin{figure}[ht]
\centering
\includegraphics[width=1.0\textwidth]{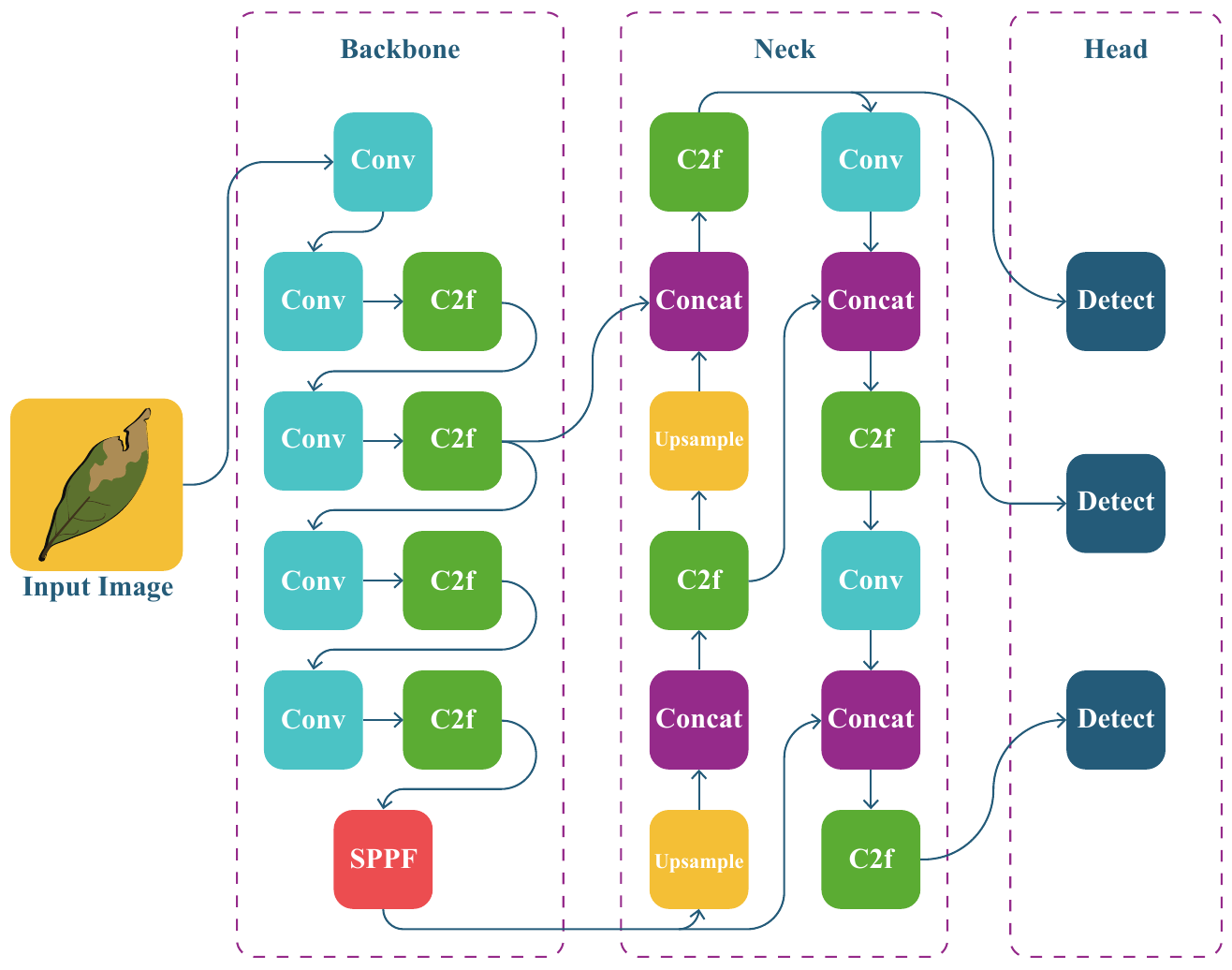}
\caption{Architecture of YOLOv8 model.}\label{fig:figure_2}
\end{figure}

\subsubsection{Models Used}\label{subsubsec:model_used_yolo}
Four YOLO architectures, such as YOLO11n, YOLO12s, YOLOv8n, and YOLOv8m, were selected for comparison. For the ability to be able to reproduce the same result with consistency, these models were trained using the Ultralytics YOLO framework.

\subsubsection{Training Configuration}\label{subsubsec:training_configuration_yolo}
The YOLO architectures use the fully convolutional network backbone for hierarchical feature extraction, along with a multi-scale detection head that predicts bounding boxes and class confidences at different resolution levels \cite{bib37}. The backbone includes residual or CSP (Cross Stage Partial) connections to improve model efficiency. The detection head uses either anchor-based or anchor-free mechanisms to simultaneously localize and classify diseases. YOLO11n and YOLO12s represent earlier and smaller-scale architectures, while YOLOv8n and YOLOv8m use newer backbone designs with improved feature pyramid networks (FPN) and decoupled heads for better accuracy.

\subsubsection{Manual Annotation for Evaluation}\label{subsubsec:manual_annotation_for_evaluation_yolo}

A manually annotated subset of 105 images (15 in each class) was generated from the test partition using Roboflow to evaluate the recognition performance. To evaluate the classification accuracy and localization accuracy across all models, this dataset provided reliable ground truth.

\subsection{RF-DETR Model}\label{subsec:RF-DETR_model}

\subsubsection{Architectural Design}\label{subsubsec:architectural_design_RF-DETR_model}

RF-DETR is an end-to-end detection architecture based on DETR that treats object detection as a direct set prediction task. It removes the need for anchor boxes and proposal generation by using a Transformer encoder--decoder to directly output bounding boxes and class labels for an image. This ``region-free'' approach is specifically designed to eliminate the need for complex components such as non-maximum suppression (NMS) and extensively hand-designed anchors \cite{bib38}.

\subsubsection{Architectural Diagram}\label{subsubsec:architectural_diagram_RF-DETR_model}

\begin{figure}[ht]
\centering
\includegraphics[width=1.0\textwidth]{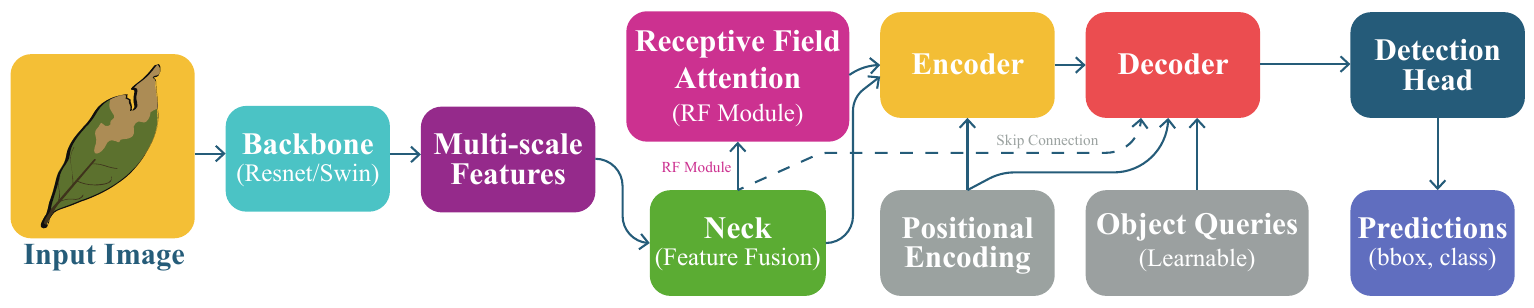}
\caption{Architecture of RF-DETR model.}\label{fig:figure_3}
\end{figure}

\subsubsection{Training Configuration}\label{subsubsec:training_configuration_RF-DETR_model}
RF-DETR typically consists of three main components:
\begin{enumerate}
    \item Backbone (Feature Extractor): A standard Convolutional Neural Network (CNN), such as ResNet (e.g., ResNet-50), is used to extract a compact set of high-level image features.

    \item Transformer Encoder-Decoder: The extracted features are flattened and passed to a Transformer encoder, which refines the feature representation. A Transformer decoder then takes a fixed set of learnable object queries and the encoded features to predict bounding boxes and classes in parallel.

    \item Prediction Head: A simple feed-forward network (FFN) that takes the decoder output and predicts the final bounding box coordinates and class probabilities for each query.

A crucial aspect of RF-DETR and DETR-like models is the Bipartite Matching Loss, which uses the Hungarian algorithm to uniquely match the predicted outputs to the ground truth objects, penalizing the difference in both class and bounding box coordinates simultaneously.

\end{enumerate}

\subsubsection{Models Used}\label{subsubsec:models_used_RF-DETR_model}
While RF-DETR is a single architecture, researchers often use different backbone networks or variations of the Transformer structure. For our comparison, the RFDETR model was trained using the default configuration, which utilizes a ResNet-50 backbone for feature extraction.

\subsection{RetinaNet Model}\label{subsec:retinaNet_model}

\subsubsection{Architectural Design}\label{subsubsec:architectural_design_retinaNet_model}
RetinaNet is a single-stage object detection model designed to address the extreme foreground-background class imbalance encountered during the training of dense detectors. It achieves this by introducing the Focal Loss, which down-weights the contribution of well-classified examples (easy negatives) and focuses training on hard, misclassified examples. The network structure is built upon standard feature pyramids, enabling robust, multi-scale object detection \cite{bib39}.

\subsubsection{Architecture Diagram}\label{subsubsec:architecture_diagram_retinaNet_model}

\begin{figure}[ht]
\centering
\includegraphics[width=1.0\textwidth]{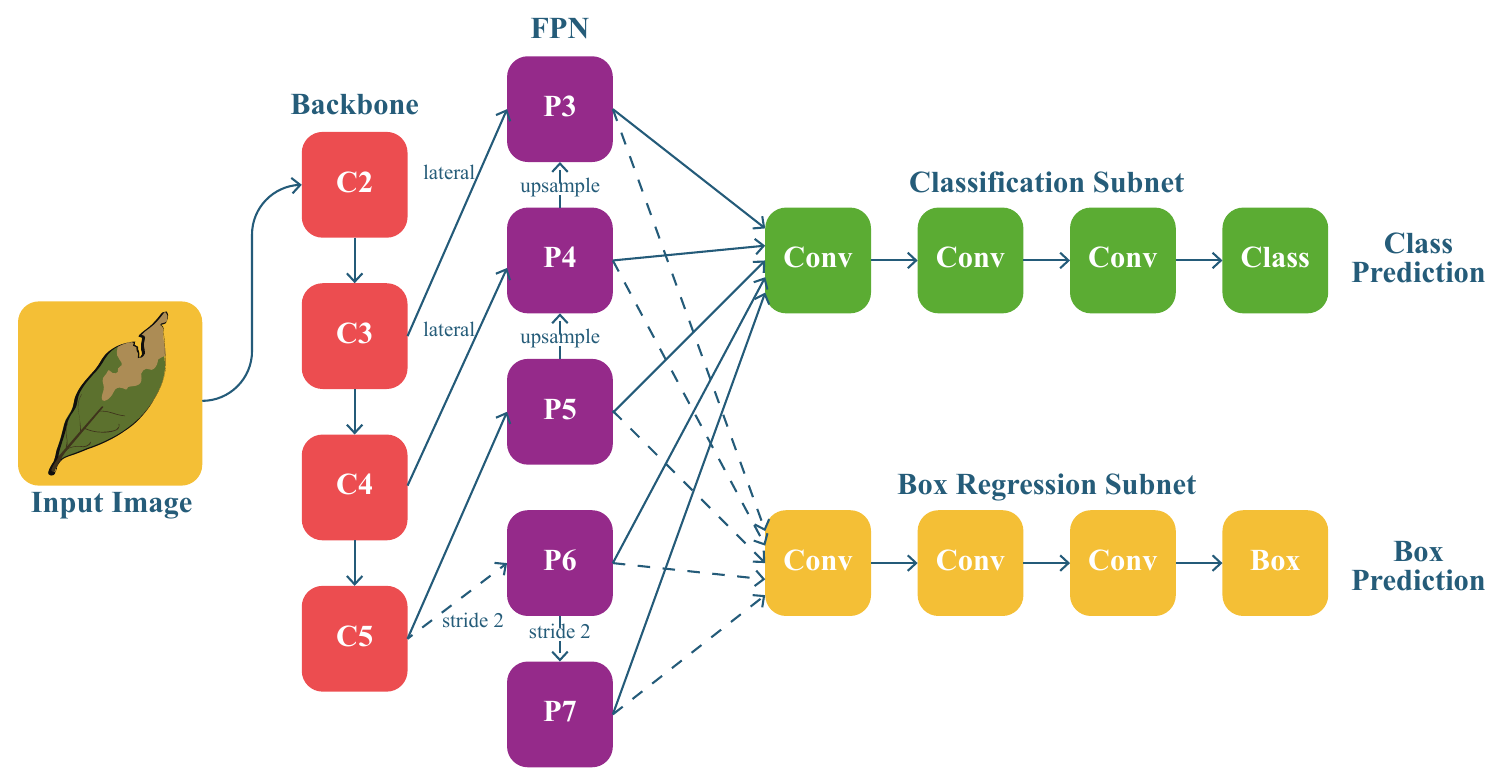}
\caption{Architecture of RetinaNet model.}\label{fig:figure_4}
\end{figure}

\subsubsection{Training Configuration}\label{subsubsec:training_configuration_retinaNet_model}
\begin{enumerate}
    \item Backbone: A standard CNN, such as ResNet-50 or ResNet-101, extracts features from the input image.
    \item Neck (Feature Pyramid Network - FPN): The FPN takes the features from the backbone's different stages and constructs a top-down pathway with lateral connections. This generates a set of semantically rich feature maps at different scales (e.g., P3, P4, P5, P6, P7), allowing the detector to find objects of various sizes effectively.
    \item Detection Heads (Classification and Regression): Two separate, small Fully Convolutional Networks (FCNs) are attached to every level of the FPN. One head performs classification (predicting the probability of K object classes at each spatial location), and the other performs bounding box regression (refining the pre-defined anchor boxes).
\end{enumerate}

\subsubsection{Models Used}\label{subsubsec:models_used_retinaNet_model}
The RetinaNet model used for comparison employed a ResNet-50 backbone integrated with a Feature Pyramid Network (FPN).

\subsection{Faster R-CNN Model}\label{subsec:faster_R-CNN_model}

\subsubsection{Architectural Design}\label{subsubsec:architectural_design_faster_R-CNN_model}
Faster R-CNN (Region-based Convolutional Neural Network) is a two-stage object detection framework and a highly influential model in computer vision. It significantly improved upon its predecessors (R-CNN and Fast R-CNN) by integrating the region proposal generation step directly into the network. The first stage generates a sparse set of high-quality object proposals, and the second stage classifies these proposals and refines their bounding boxes. This architecture is known for its high accuracy \cite{bib40}.

\subsubsection{Architectural Diagram}\label{subsubsec:architectural_diagram_faster_R-CNN_model}

\begin{figure}[ht]
\centering
\includegraphics[width=1.0\textwidth]{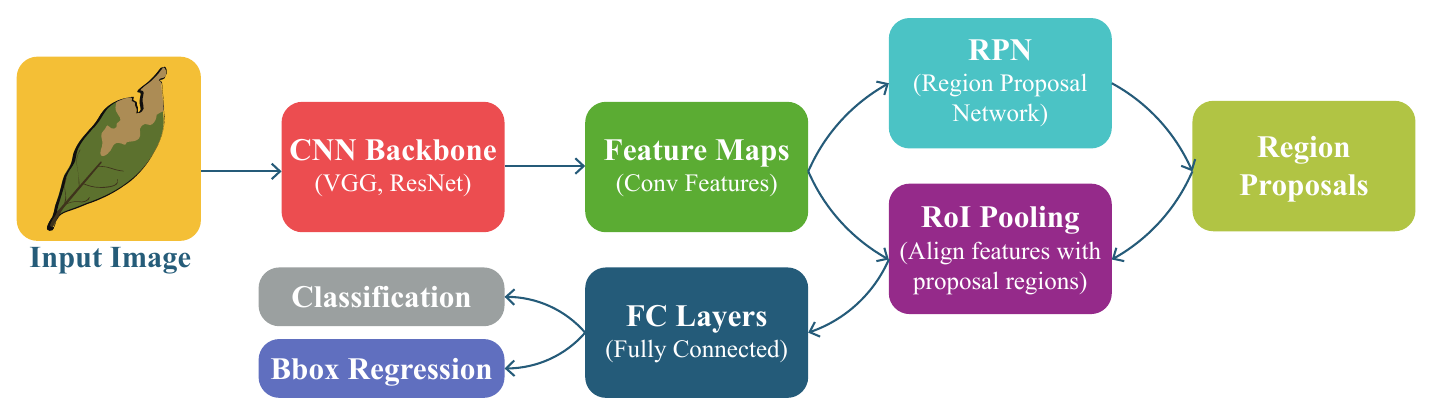}
\caption{Architecture of Faster R-CNN model.}\label{fig:figure_5}
\end{figure}

\subsubsection{Training Configuration}\label{subsubsec:training_configuration_faster_R-CNN_model}
The architecture consists of four key modules:
\begin{enumerate}
    \item Backbone (Feature Extractor): A standard CNN, like ResNet-50 or VGG, is used to extract the shared feature map from the input image.
    \item Region Proposal Network (RPN): This is the first stage. A small network slides over the feature map and simultaneously predicts:\begin{itemize}
        \item Objectiveness Score: The probability that a region contains an object (foreground/background classification).
        \item Bounding Box Regression: Offsets to refine the coordinates of pre-defined anchor boxes.
        \item The RPN generates a few thousand candidate object proposals.
    \end{itemize}
    \item RoI Pooling/Align: The high-scoring proposals from the RPN are mapped back to the feature map, and a fixed-size feature vector is extracted for each proposal using RoI (Region of Interest) Pooling or the more precise RoI Align layer.
    \item Detection Heads (Classification and Regression): These are the final layers that take the fixed-size feature vector for each proposal and perform the final object classification (into K specific classes) and a second, more precise bounding box regression.
\end{enumerate}

\subsubsection{Models Used}\label{subsubsec:models_used_faster_R-CNN_model}

The Faster R-CNN model was implemented with a ResNet-50 backbone and enhanced with a Feature Pyramid Network (FPN) to improve multi-scale detection performance.

\subsection{Custom CNN: LightTeaNet}\label{subsec:lightTeaNet}
For multi-label tea leaf disease classification and localization, LightTeaNet, a lightweight CNN, has been proposed. The architecture combines channel attention mechanisms to enhance feature discriminability with Depthwise Separable Convolutions (DWConvBlock) \cite{bib10,bib35} for efficient parameter utilization.

\subsubsection{Architectural Design}\label{subsubsec:architectural_design_lightTeaNet}
LightTeaNet follows a compact, attention-oriented design to balance accuracy and low computational cost. Its architecture focuses on efficiently extracting features and interpreting them via activation-based localization. Key modules:
\begin{itemize}
    \item DWConvBlock: Performs spatial and channel-wise convolutions separately, reducing the number of parameters.
    \item Channel focus \cite{bib4,bib18,bib30}: Highlights disease-related channels by re-adapting functional maps.
    \item Global Average Pooling (GAP): Converts spatial features into compact global descriptions.
    \item Fully connected layer: Outputs seven sigmoid-enabled nodes representing disease probabilities.
\end{itemize}

\subsubsection{Architectural Diagram}\label{subsubsec:architectural_diagram_lightTeaNet}

\begin{figure}[ht]
\centering
\includegraphics[width=1.0\textwidth]{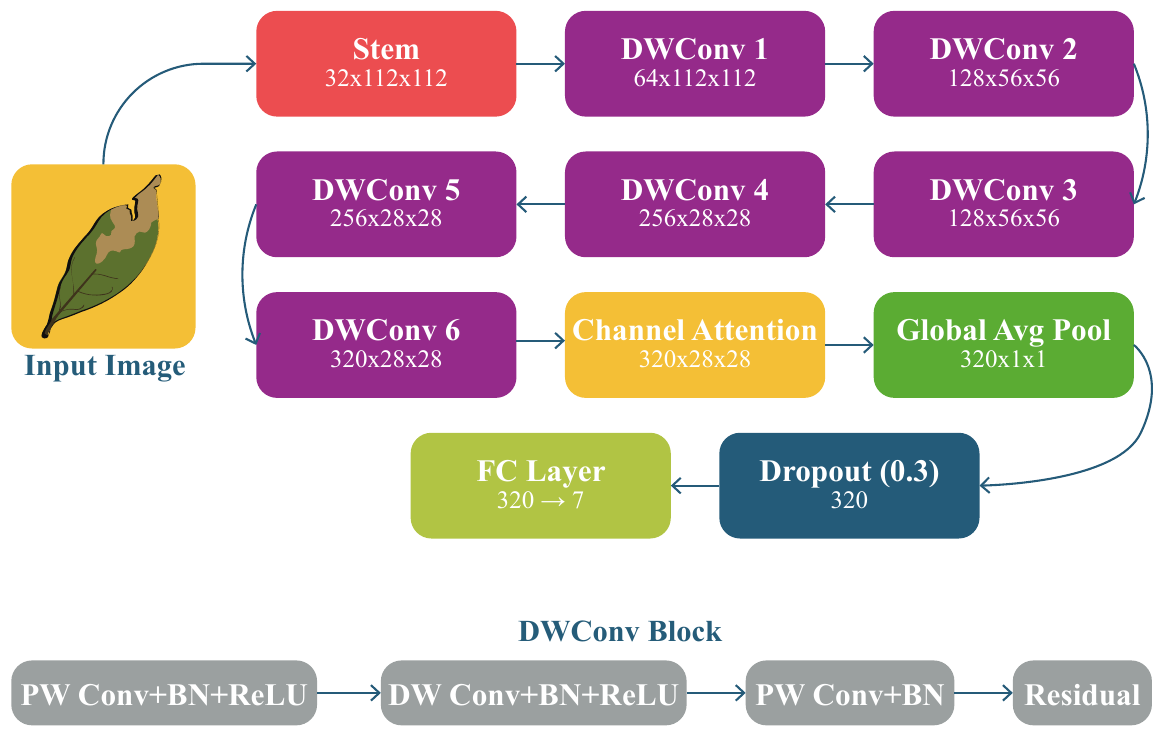}
\caption{Architecture of the proposed LightTeaNet model.}\label{fig:figure_6}
\end{figure}

\subsubsection{Training Configuration}\label{subsubsec:training_configuration_lightTeaNet}
The training configuration used binary cross-entropy as the loss function and Adam optimizer with a fixed learning rate of 0.001. Training was performed with a batch size of 16 for 300 epochs. To increase efficiency and reduce GPU memory usage, mixed precision training was enabled using Autocast and Gradscaler. The model implemented early stopping, and checkpoints were saved whenever an improvement was observed in the best weighted F1 score, which served as the model's checkpoint criteria.

\subsubsection{Class Activation Mapping (CAM) for Localization}\label{subsubsec:class_activation_mapping_(CAM)_for_localization_lightTeaNet}
The class activation mapping (CAM) method \cite{bib8,bib15,bib28} was used to visualize and extract disease-affected regions from the final functional map. The final attention layer (Features.6) was used to compute class-specific activation maps using the weights of the last fully connected layer. The generated heatmap was processed using:
\begin{itemize}
    \item Threshold to isolate high response regions.
    \item Morphological operations to remove noise and increase shape stability.
    \item Bounding box extraction for accurate detection of diseased areas.

These bounding boxes act as automatic annotations for training YOLO models, making LightNet a self-supervised labelling mechanism.    
    
\end{itemize}

\subsection{Evaluation}\label{subsec:evaluation}

We tested all five models---LightNet, YOLO11n, YOLO12s, YOLOv8n, and YOLOv8m---against a set of 105 images that we'd manually labeled ourselves. We looked at how well they classified things (checking accuracy, F1 scores, and Hamming loss) and how precisely they located objects (using IOU and mAP measurements). When we compared everything, LightTenet really stood out. It did a solid job automating the annotation process and managed to identify things pretty accurately without needing much manual work from us.

In summary, the proposed methodology combines efficient dataset preparation, a lightweight CNN architecture, and weakly supervised learning through Class Activation Mapping (CAM). LightTeaNet creates its own pseudo-annotations and then we pit it against the YOLO models to see how it performs. What's interesting is that it shows weak supervision can actually give you reliable results---no manual labeling needed. In the next section, we'll dive into what happened during the experiments and break down the results.

Finally, all ten models---LightTeaNet, YOLO11n, YOLO12s, YOLOv8n, YOLOv8m, RF-DETR, DETR, RetinaNet, and Faster R-CNN---were evaluated using a manually annotated test set of 105 images. Both classification metrics (accuracy, F1 score, Hamming loss) and localization metrics (IoU, mAP) were analyzed. When we compared everything, it became clear that LightTeaNet really shines at automated annotation. It manages to match the identification accuracy of top-tier object detection models without needing much hands-on work from humans. That's pretty impressive when you stack it up against all the state-of-the-art architectures out there.

To sum it up, our approach brings together three key things: efficient dataset prep, a lightweight CNN design, and weakly supervised learning using Class Activation Mapping (CAM). LightTeaNet generates its own pseudo-annotations and we tested it against a whole range of object detection models. What we found is that weak supervision can give you solid, reliable results without anyone having to manually label data. The following section presents the experimental outcomes and analyzes the results in detail.

\section{Result}\label{sec:result}
To evaluate the effectiveness of the proposed LightTeaNet, a series of experiments were conducted comparing it with several YOLO architectures. The analysis focuses on both classification performance and localization capability, providing a clear understanding of how the model performs under weak supervision. The following results highlight LightTeaNet's balance between accuracy, efficiency, and interpretability.

\subsection{Overview}\label{subsec:result_overview}
Four YOLO variants were assessed and compared with the performance of LightTeaNet, using the dataset. The same 105 manually annotated test images, which contain 114 bounding boxes across seven classes, were used for testing all models. Each model was evaluated in terms of classification and localization performance.

\subsection{Classification Results}\label{subsec:classification_results}

The classification performance of all models was evaluated using Precision (P), Recall (R), and F1-Score, computed over the test set. As summarized in Table~\ref{tab:tab1}, the proposed LightTeaNet achieved the highest overall classification accuracy, followed by incremental improvements across YOLO variants from v8n to v12s.

\begin{figure}[ht]
\centering
\includegraphics[width=1.0\textwidth]{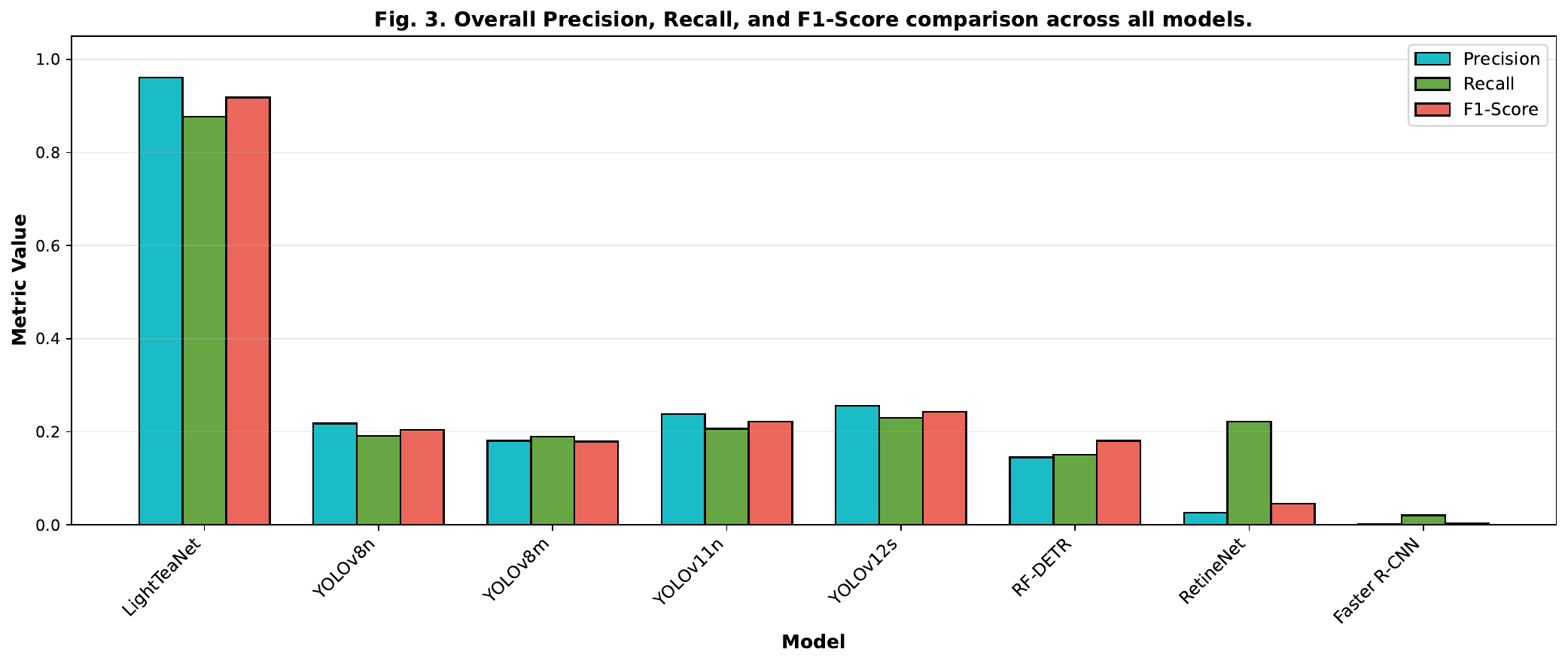}
\caption{Overall Precision, Recall, and F1-Score comparison across all models.}\label{fig:figure_7}
\end{figure}

\begin{figure}[ht]
\centering
\includegraphics[width=1.0\textwidth]{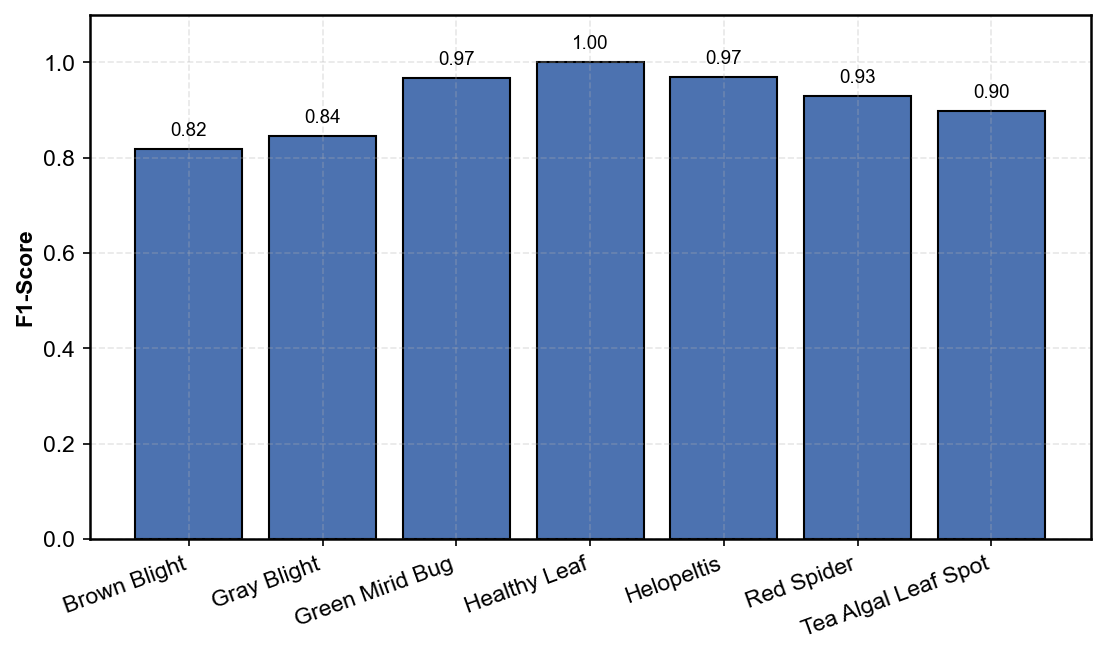}
\caption{Per-class F1-Score distribution of LightTeaNet.}\label{fig:figure_8}
\end{figure}

Per-class classification performance of the proposed CNN is reported in Table~\ref{tab:tab2}, showing consistently high accuracy across all seven disease categories. The bar graph in Figure~\ref{fig:figure_7} presents the overall classification metrics for all models \cite{bib24}, while Figure~\ref{fig:figure_8} visualizes the per-class F1-Scores of LightTeaNet.

\begin{table}[ht]
\centering
\caption{Overall classification performance comparison among all models}
\label{tab:tab1}
\begin{tabular}{@{}llll@{}}
\hline
Model & Precision (P) & Recall (R) & F1-Score \\
\hline
LightTeaNet & 0.9615 & 0.8772 & 0.9179 \\
YOLOv8n & 0.218 & 0.191 & 0.204 \\
YOLOv8m & 0.181 & 0.190 & 0.179 \\
YOLO11n & 0.238 & 0.207 & 0.222 \\
YOLO12s & 0.256 & 0.230 & 0.243 \\
RF-DETR & 0.145 & 0.151 & 0.181 \\
RetinaNet (ResNet-50-FPN) & 0.026 & 0.222 & 0.0461 \\
Faster R-CNN (ResNet-50-FPN) & 0.0016 & 0.0204 & 0.003 \\
\hline
\end{tabular}
\end{table}

\begin{table}[ht]
\centering
\caption{Per-class classification results of LightTeaNet}
\label{tab:tab2}
\begin{tabular}{@{}llll@{}}
\hline
Class Name & Precision & Recall & F1-Score \\
\hline
Brown Blight & 0.9000 & 0.7500 & 0.8182 \\
Gray Blight & 0.9048 & 0.7917 & 0.8444 \\
Green Mirid Bug & 1.0000 & 0.9375 & 0.9677 \\
Healthy Leaf & 1.0000 & 1.0000 & 1.0000 \\
Helopeltis & 1.0000 & 0.9412 & 0.9697 \\
Red Spider & 1.0000 & 0.8667 & 0.9286 \\
Tea Algal Leaf Spot & 0.9286 & 0.8667 & 0.8966 \\
\hline
\end{tabular}

\end{table}

\subsection{Localization Results}\label{subsec:localization_results}

\begin{figure}[ht]
\centering
\includegraphics[width=1.0\textwidth]{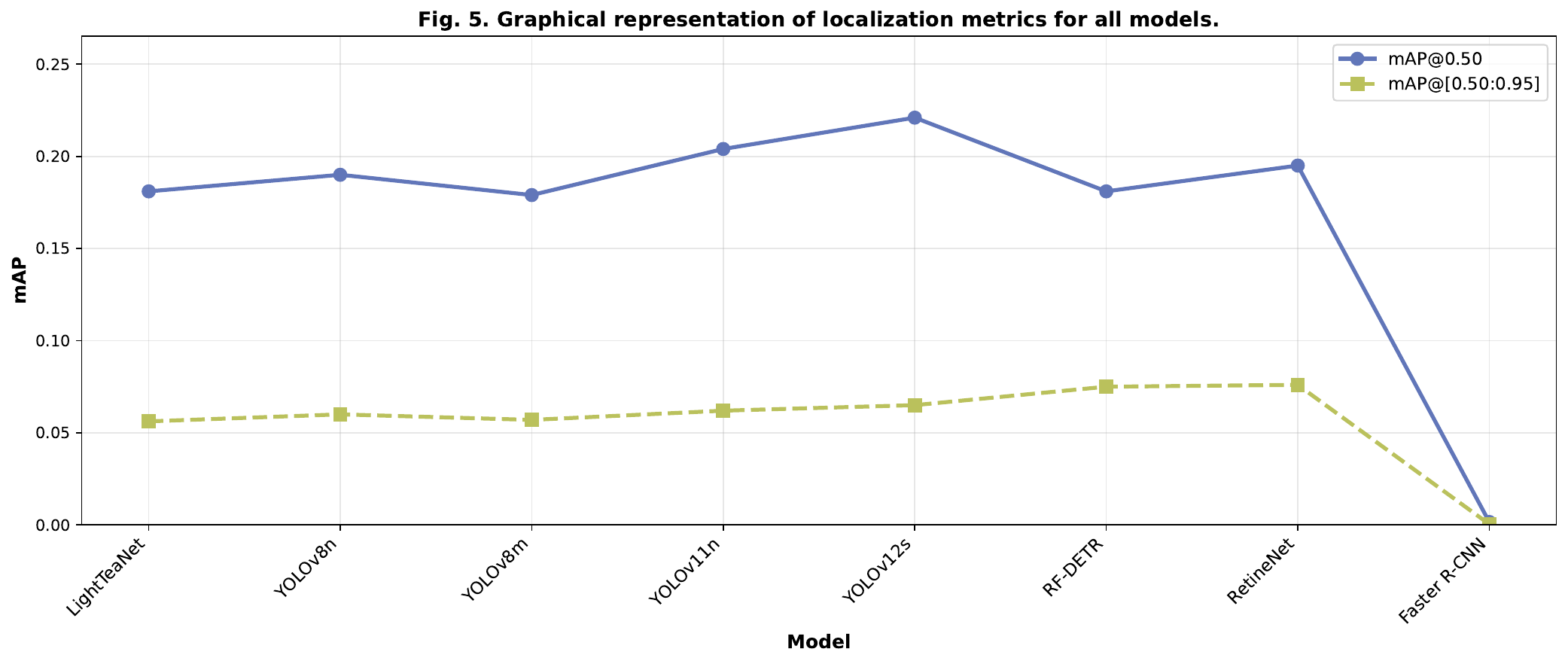}
\caption{Graphical representation of localization metrics for all models.}\label{fig:figure_9}
\end{figure}

By comparing each model’s detected or highlighted disease regions against manual bounding boxes, the localization ability was assessed using Mean Average Precision (mAP) at IoU thresholds of 0.50 and 0.50:0.95. The YOLO models reported in Table~\ref{tab:tab3} were trained using our CAM-generated pseudo-annotations rather than manual bounding box annotations. Table~\ref{tab:tab3} presents the quantitative localization results, while the graphical comparison of localization metrics is shown in Figure~\ref{fig:figure_9}.

\begin{table}[ht]
\centering
\caption{Localization performance (mAP@0.50 and mAP@[0.50:0.95]) comparison across all models}
\label{tab:tab3}
\begin{tabular}{@{}lll@{}}
\hline
Model & mAP@0.50 & mAP@[0.50:0.95] \\
\hline
LightTeaNet & 0.1810 & 0.0562 \\
YOLOv8n & 0.190 & 0.060 \\
YOLOv8m & 0.179 & 0.057 \\
YOLO11n & 0.204 & 0.062 \\
YOLO12s & 0.221 & 0.065 \\
RF-DETR & 0.181 & 0.075 \\
RetinaNet (ResNet-50-FPN) & 0.195 & 0.076 \\
Faster R-CNN (ResNet-50-FPN) & 0.0016 & 0.0004 \\
\hline
\end{tabular}
\end{table}

\subsection{Qualitative Visualization}\label{subsec:qualitative_visualization}
\begin{figure}[ht]
\centering
\includegraphics[width=1.0\textwidth]{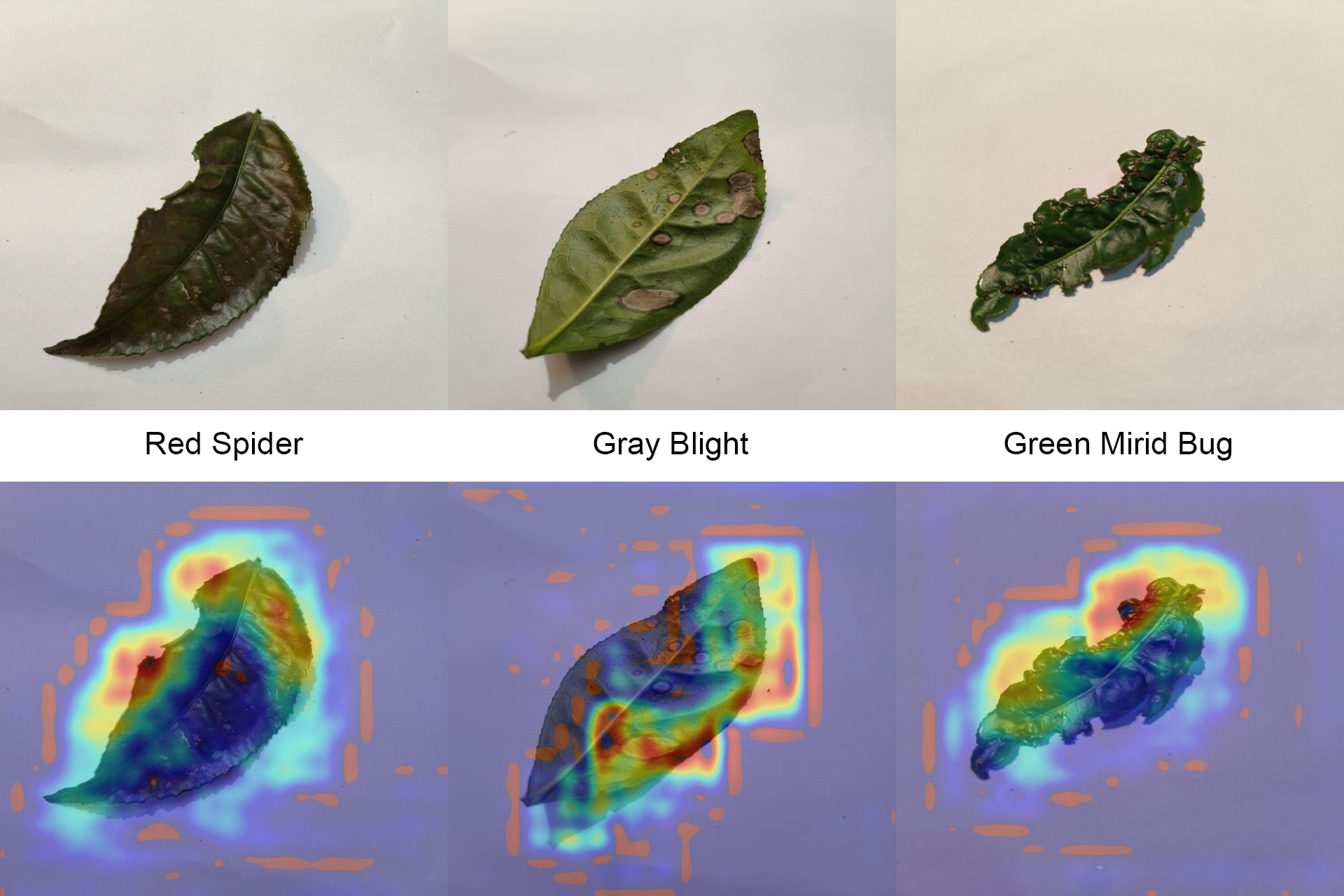}
\caption{GradCAM visualizations showing disease localization (activation maps) of LightTeaNet.}\label{fig:figure_10}
\end{figure}

Figure~\ref{fig:figure_10} presents representative examples of GradCAM outputs generated by LightTeaNet, showing accurate disease localization regions that closely align with the manually annotated bounding boxes. The Infected leaf regions were successfully highlighted by the model without explicit bounding box training, demonstrating the potential of the model for weakly supervised localization.

\subsection{Summary of Quantitative Findings}\label{subsec:summary_of_quantitative_findings}

A consolidated summary of both classification and localization metrics for all models is presented in Table~\ref{tab:tab4}, providing an integrated overview of model performance across both tasks.

\begin{table}[ht]
\caption{Summary of combined classification and localization performance for all models}
\centering
\label{tab:tab4}
\begin{tabular}{@{}llllll@{}}
\toprule
Model & Precision (P) & Recall (R) & F1-Score & mAP@0.50 & mAP@[0.50:0.95] \\
\midrule
LightTeaNet              & 0.9615 & 0.8772 & 0.9179 & 0.181 & 0.0562 \\
YOLOv8n                  & 0.218  & 0.191  & 0.204  & 0.190 & 0.060  \\
YOLOv8m                  & 0.181  & 0.190  & 0.179  & 0.179 & 0.057  \\
YOLO11n                  & 0.238  & 0.207  & 0.222  & 0.204 & 0.062  \\
YOLO12s                  & 0.256  & 0.230  & 0.243  & 0.221 & 0.065  \\
RF-DETR                  & 0.145  & 0.151  & 0.181  & 0.181 & 0.075  \\
RetinaNet (ResNet-50-FPN)& 0.026  & 0.222  & 0.0461 & 0.195 & 0.076  \\
Faster R-CNN (ResNet-50-FPN)
                         & 0.0016 & 0.0204 & 0.003  & 0.0016 & 0.0004 \\
\bottomrule
\end{tabular}
\end{table}

\begin{figure}[ht]
\centering
\includegraphics[width=1.0\textwidth]{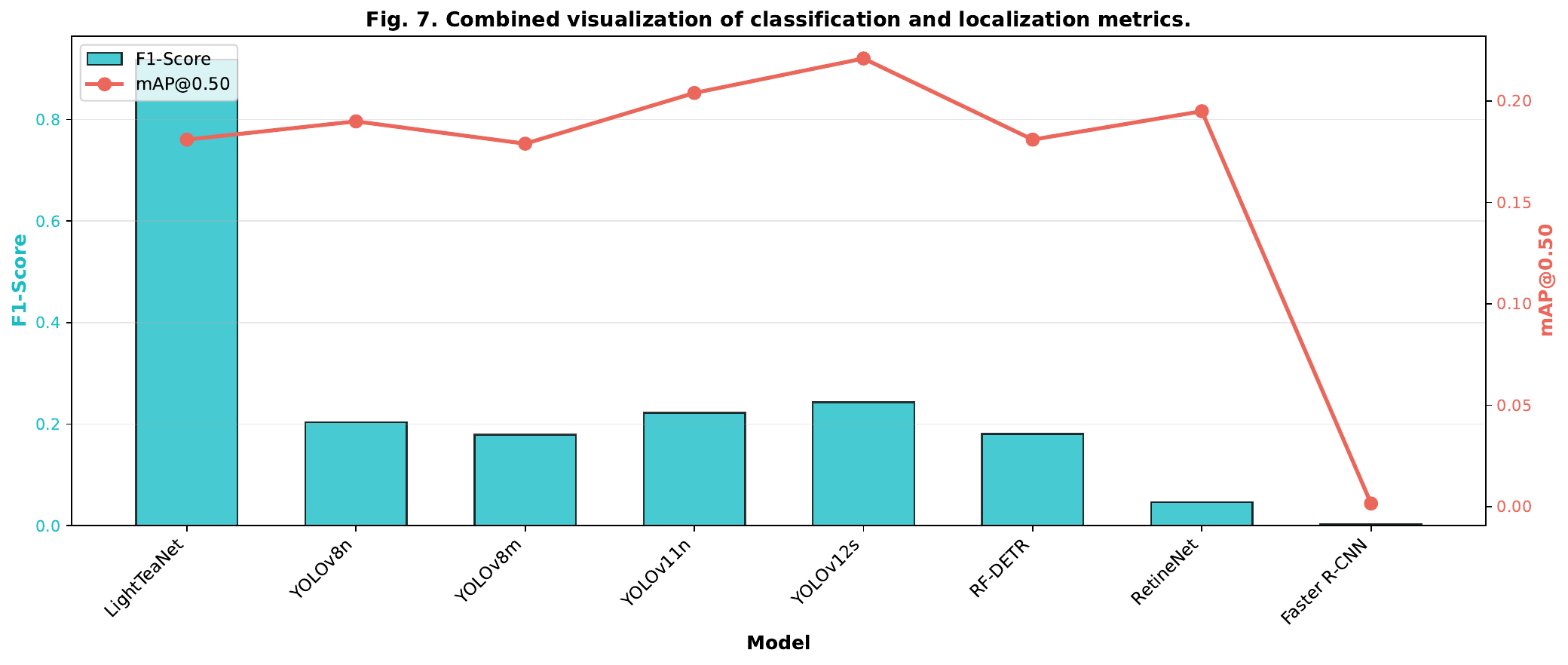}
\caption{Combined visualization of classification and localization metrics across all models.}\label{fig:figure_11}
\end{figure}

Looking at everything together, LightTeaNet actually outperforms the fully supervised YOLO models when it comes to classification, and it holds its own pretty well on localization too. What's really cool is that it doesn't need any manual annotations---it figures out where the disease spots are using its CAM-based visualization. These results show that LightTeaNet could be a genuinely practical and lightweight tool for monitoring crop diseases in the real world.

\subsection{Statistical Validation of Model Performance}\label{subsec:statistical_validation_of_model_performance}

To check whether the performance differences between LightTeaNet and seven other models were statistically meaningful, we used a permutation test with 10,000 permutations. For each comparison, the per-image scores from both models were combined and randomly shuffled into two groups of the same sizes as the originals. The p-value was then calculated as the share of shuffled differences that were as large as or larger than the actual observed difference. This method builds a proper null distribution centred at zero, which allows valid statistical testing under the assumption that both models perform equally. We also measured effect sizes using Cohen's d and calculated 95\% confidence intervals using 10,000 bootstrap resamples. A result was considered statistically significant if $p < 0.05$.

LightTeaNet achieved a mean mAP@0.50 of 0.181 on the test set. The permutation tests showed that LightTeaNet's mAP was not significantly different from six of the seven baselines, namely YOLOv8n, YOLOv8m, YOLOv11n, YOLOv12s, RF-DETR, and RetinaNet (all $p > 0.05$, with negligible to small effect sizes), meaning their localisation performance was broadly comparable. Faster R-CNN was the only model with a significantly higher mAP (mean = 0.344; difference = $-0.163$, 95\% CI [$-0.275$, $-0.050$], $p = 0.005$, $d = -0.39$, small effect). However, LightTeaNet significantly outperformed all seven baselines on Precision, Recall, and F1-Score ($p < 0.0001$ in all cases), with large effect sizes (Cohen's d between 1.19 and 4.05). This shows that LightTeaNet is a much stronger model for classifying disease presence, even though its raw localisation score was lower than Faster R-CNN.

\section{Comparative Study}\label{sec:comparative_study}
To evaluate LightTeaNet against existing CNN-based approaches, we recreated two architectures proposed by Das et al.\ \cite{bib41} and Shikdar et al.\ \cite{bib42} and evaluated them on our dataset under identical experimental conditions. Because the original studies were conducted on different datasets, directly comparing published metrics would be methodologically unsound; re-implementing their architectures on our data provides a more meaningful basis for comparison.

Das et al.'s \cite{bib41} lightweight sequential CNN struggled considerably on our dataset, yielding only 60.00\% accuracy with a precision of 61.31\%, recall of 60.00\%, and F1-score of 59.24\%. Per-class inspection revealed that the model had particular difficulty distinguishing Tea Algal Leaf Spot from Red Spider, suggesting that its limited depth and representational capacity are inadequate for capturing the fine-grained inter-class variations present in our seven-class dataset. Shikdar et al.'s \cite{bib42} deeper architecture, which incorporates Batch Normalization and Global Average Pooling, fared considerably better, reaching 92.38\% accuracy and a weighted F1-score of 92.41\%, with perfect classification on the Healthy Leaf and Red Spider categories.

LightTeaNet substantially outperforms Das et al.'s model and remains highly competitive with Shikdar et al.'s architecture, achieving a precision of 96.15\%, recall of 87.72\%, and F1-score of 91.79\%. A critical distinction, however, is that Shikdar et al.'s model is a classification-only system with no spatial localization capability. LightTeaNet, by contrast, addresses both disease classification and lesion localization within a single unified framework, with its detection performance further quantified through mAP metrics. Statistical analysis confirms that LightTeaNet's classification performance significantly exceeds all seven object detection baselines evaluated in this study ($p < 0.0001$, Cohen's d ranging from 1.19 to 4.05). Taken together, these results demonstrate that LightTeaNet offers a more complete and practically deployable solution for tea leaf disease detection, without meaningfully sacrificing classification accuracy relative to specialized classification models.

\begin{table}[ht]
\centering
\caption{Comparative classification performance of LightTeaNet against CNN-based approaches}
\label{tab:comparative}
\begin{tabular}{@{}lllll@{}}
\hline
Model & Accuracy & Precision (P) & Recall (R) & F1-Score \\
\hline
Das et al.\ \cite{bib41} & 0.6000 & 0.6131 & 0.6000 & 0.5924 \\
Shikdar et al.\ \cite{bib42} & 0.9238 & --- & --- & 0.9241 \\
\textbf{LightTeaNet (Ours)} & --- & \textbf{0.9615} & \textbf{0.8772} & \textbf{0.9179} \\
\hline
\end{tabular}
\end{table}

\section{Discussion}\label{sec:discussion}
The advantages of the proposed LightTeaNet framework over YOLO-based detection models are clearly demonstrated in the comparative experimental results. LightTeaNet achieved the highest classification performance with Precision = 0.9615, Recall = 0.8772, and F1-Score = 0.9179. This indicates that the model effectively captures the subtle spatial and textural characteristics of diseased tea leaves from image-level supervision alone. The depthwise separable design guaranteed computational efficiency appropriate for field deployment on low-resource devices \cite{bib10,bib35}, while the integration of channel attention \cite{bib4,bib17,bib18,bib19} enhanced sensitivity to disease-relevant feature representations.

It is important to explicitly note that the classification comparison between LightTeaNet and YOLO-based detectors is not strictly equivalent in terms of evaluation mechanics. LightTeaNet is optimized and evaluated purely for image-level classification, whereas YOLO models are trained as object detectors and are jointly optimized for classification, bounding box regression, and objectness estimation. Consequently, YOLO classification metrics are inherently influenced by localization and regression errors, which may significantly penalize their reported F1-scores. Therefore, the superior classification F1-score of LightTeaNet should be interpreted in the context of architectural and objective-function differences rather than as a direct architectural superiority in pure classification alone.

Regarding localization, LightTeaNet’s CAM-based visual maps achieved mAP@0.50 = 0.1810, which is reasonably close to the best fully supervised YOLO model result (mAP@0.50 = 0.221). Although CAM outputs are intrinsically coarser than explicitly trained detection heads, their agreement with manual annotations highlights the model’s inherent spatial awareness. More importantly, the CAM outputs enable automatic generation of pseudo-bounding boxes, which can be used to bootstrap fully supervised object detection training under weak supervision.

Additionally, LightTeaNet provides interpretability benefits that conventional object detectors frequently lack. The Grad-CAM visualizations \cite{bib6,bib26,bib31} offer intuitive explanations of the model’s predictions, increasing transparency and trust in AI-driven agricultural decision-support systems. This interpretability is particularly valuable for adoption by farmers and agronomists.

Nevertheless, several limitations remain. The coarse granularity of CAM-based localization constrains fine-scale lesion boundary delineation. Furthermore, real-world deployment may be affected by variations in illumination, leaf orientation, and background clutter. Future research could explore spatial refinement modules, multi-scale attention mechanisms \cite{bib17,bib19}, and semi-supervised learning strategies \cite{bib21,bib29} to further enhance localization precision and domain robustness.

\section{Conclusion}\label{sec:conclusion}

LightTeaNet, a lightweight, weakly supervised CNN for multi-label tea leaf disease detection and localization, was presented in this work. The model outperformed a broad range of fully supervised object detection architectures (including YOLO variants, Transformer-based DETR, and two-stage/single-stage detectors) in terms of classification performance and achieved competitive localization accuracy when trained exclusively on image-level labels. LightTeaNet eliminates the need for human labeling by using Class Activation Mapping for self-supervised annotation creation, which drastically cuts down on computational overhead and dataset preparation time.

The model is ideal for real-time agricultural monitoring and deployment on devices with limited resources because of its robust classification capabilities, interpretability using GradCAM visualization, and lightweight construction. Additionally, the underlying architecture supports scalable and economical smart farming applications by being generalizable to different crop disease detection tasks. In order to measure plant health holistically, future research will concentrate on expanding the technique to multimodal data sources and improving localization granularity through improved attention processes.

\bmhead{Acknowledgements}

The authors acknowledge the use of Claude (Anthropic) for assistance with English language polishing and grammar correction. All scientific content, methodology, experimental design, analysis, and conclusions presented in this manuscript are solely the responsibility of the authors.

\textit{This version of the article has been accepted for publication, after peer review, but is not the Version of Record and does not reflect post-acceptance improvements or any corrections. The Version of Record is available online at: \url{https://doi.org/10.1007/s00521-026-12329-z}. Use of this Accepted Version is subject to the publisher's Accepted Manuscript terms of use \url{https://www.springernature.com/gp/open-research/policies/accepted-manuscript-terms}.}

\bibliography{sn-bibliography}% common bib file

\end{document}